\documentclass[conference]{IEEEtran}
\IEEEoverridecommandlockouts

\usepackage{cite}
\usepackage{amsmath,amssymb,amsfonts}
\usepackage{textcomp}
\usepackage{xcolor}
\usepackage{algorithmic}
\usepackage{graphicx}
\usepackage{algorithm,algorithmic}
\usepackage{textcomp}
\usepackage[hidelinks]{hyperref}
\usepackage{booktabs}
\usepackage{subfigure}

\def\BibTeX{{\rm B\kern-.05em{\sc i\kern-.025em b}\kern-.08em
    T\kern-.1667em\lower.7ex\hbox{E}\kern-.125emX}}

\begin{document}

\title{AZT1D: A Real-World Dataset for Type 1 Diabetes}
\author{Saman Khamesian$^{1, 2}$, Asiful Arefeen$^{1, 2}$, Bithika M. Thompson$^{3}$, Maria Adela Grando$^{1}$, and Hassan Ghasemzadeh$^{1}$
\thanks{$^{1}$College of Health Solutions, Arizona State University, Phoenix, AZ, USA.}
\thanks{$^{2}$School of Computing and Augmented Intelligence, Arizona State University, Tempe, AZ, USA.}
\thanks{$^{3}$Department of Endocrinology, Mayo Clinic Arizona, Scottsdale, AZ, USA.}
\thanks{{Corresponding author: \textcolor{blue}{skhamesi@asu.edu}}}
\thanks{{Dataset link: \href{https://data.mendeley.com/datasets/gk9m674wcx/1}{\textcolor{blue}{10.17632/gk9m674wcx.1}}}}
}

\maketitle

\begin{abstract}
High-quality real-world datasets are essential for advancing data-driven approaches in type 1 diabetes (T1D) management, including personalized therapy design, digital twin systems, and glucose prediction models. However, progress in this area has been limited by the scarcity of publicly available datasets that offer detailed and comprehensive patient data. To address this gap, we present $AZT1D$, a dataset containing data collected from 25 individuals with T1D on automated insulin delivery (AID) systems. AZT1D includes continuous glucose monitoring (CGM) data, insulin pump and insulin administration data, carbohydrate intake, and device mode (regular, sleep, and exercise) obtained over 6–8 weeks for each patient. Notably, the dataset provides granular details on bolus insulin delivery (i.e., total dose, bolus type, correction-specific amounts) features that are rarely found in existing datasets. By offering rich, naturalistic data, AZT1D supports a wide range of artificial intelligence and machine learning applications aimed at improving clinical decision-making and individualized care in T1D.
\end{abstract}

\begin{IEEEkeywords}
e-health, wearable sensors, type 1 diabetes, continuous glucose monitoring, automated insulin delivery systems, machine learning.
\end{IEEEkeywords}

\section{Introduction}
Type 1 diabetes (T1D) is a chronic autoimmune disease characterized by the destruction of pancreatic beta cells, resulting in an absolute insulin deficiency and a lifelong dependency on exogenous insulin therapy. Effective management of T1D requires continuous monitoring and precise insulin administration to maintain blood glucose levels within a target range, as poor glycemic control increases the risk of severe complications, including cardiovascular disease, neuropathy, and kidney failure \cite{atkinson2014type, diabetes2005intensive}. To mitigate these risks and improve the quality of life for individuals with T1D, continuous glucose monitoring (CGM) systems and automated insulin delivery (AID) systems have been developed \cite{desalvo2013continuous, limbert2024automated}. CGM devices, such as the Dexcom G6 and G7, provide real-time glucose readings, enabling patients and healthcare providers to monitor glycemic trends and respond to fluctuations more effectively. AID systems integrate CGM data with insulin pumps to automate insulin delivery, reducing the burden of manual insulin adjustments and enhancing glucose regulation. The availability of real-world patient data from these advanced technologies is essential for developing more effective diabetes management strategies.

Many machine learning approaches have been proposed to improve T1D care, such as therapy optimization, digital twin modeling, and counterfactual analysis. However, the development and evaluation of these methods are often limited by the lack of comprehensive real-world datasets. Most publicly accessible datasets are either simulated or drawn from small cohorts, reducing their generalizability to broader patient populations \cite{marling2020ohiot1dm, hidalgo2024hupa}.

To address this gap, we introduce \textit{AZT1D}, a real-world dataset collected from 25 individuals with T1D using AID systems during routine clinical care. In addition to CGM data, AZT1D includes detailed insulin administration logs, carbohydrate intake, and contextual device mode information (regular, sleep, exercise). Notably, AZT1D provides fine-grained bolus-related variables, including total bolus dose, bolus type (standard, correction, or automatic), and the portion of insulin specifically used for correction. To the best of our knowledge, this is the only publicly available dataset that offers this level of detail on bolus insulin events. By capturing naturalistic, multi-week T1D management behavior, AZT1D enables a wide range of applications, from interpretable modeling and treatment recommendation to simulation-based evaluation frameworks.

\begin{figure*}[t]
    \centering
    \subfigure[Basal rate extraction]{
        \includegraphics[width=\textwidth]{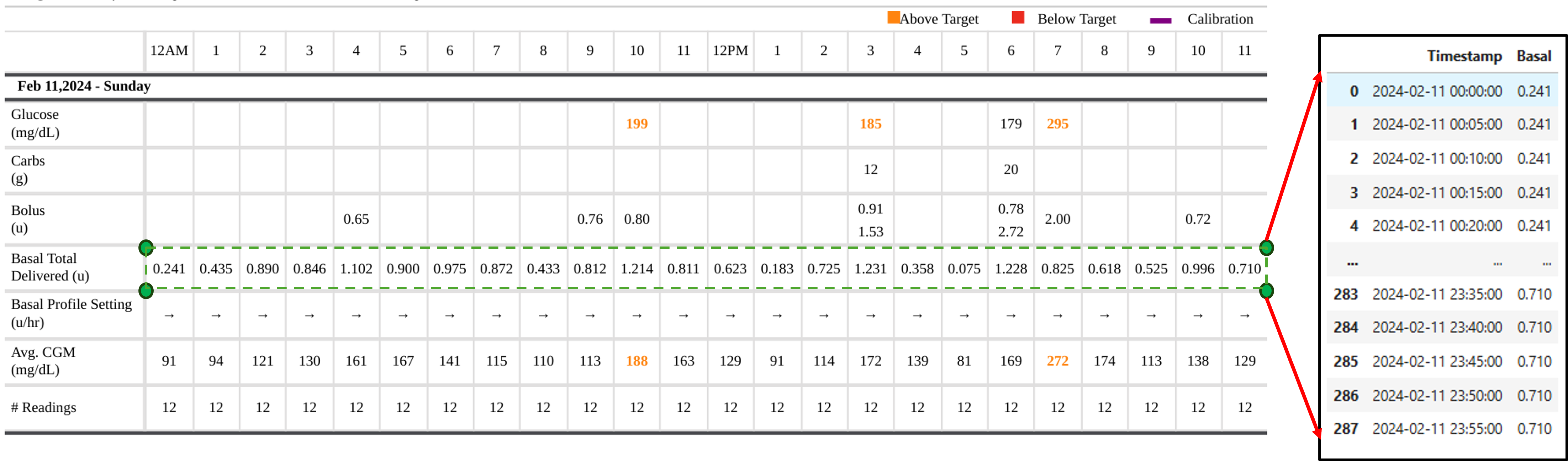}
        \label{basal}
    }
    \subfigure[AID device mode extraction]{
        \includegraphics[width=\textwidth]{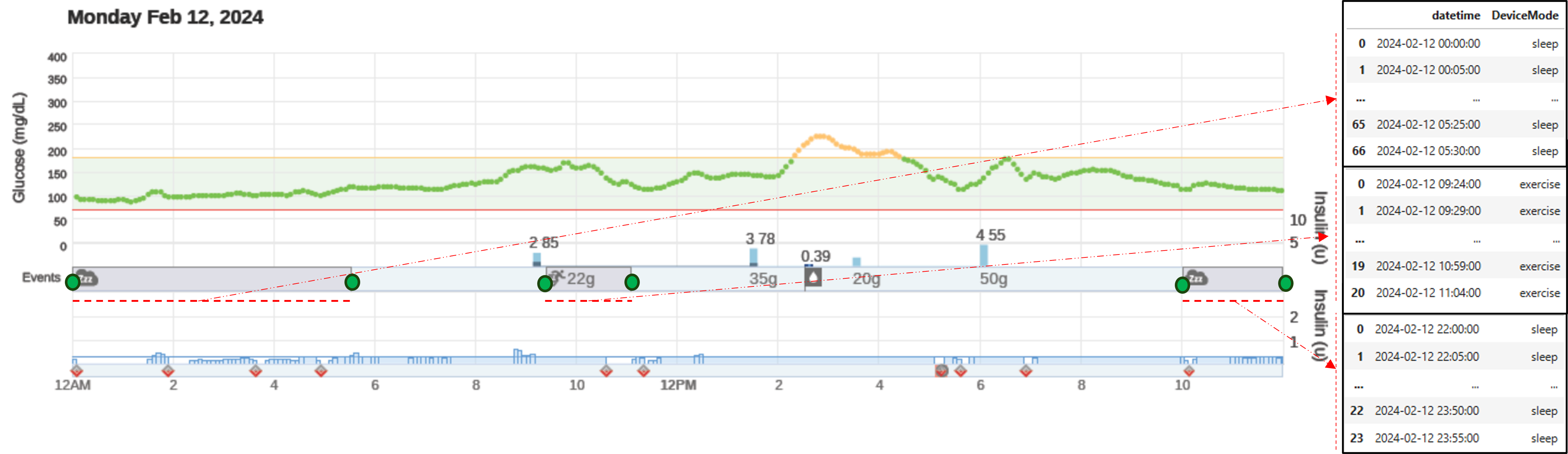}
        \label{mode}
    }
    \caption{Basal rate and the device mode extraction from the PDFs using OCR and coordinate system.}
    \label{basal_modes}
\end{figure*}

\section{Study Design and Participants}
This study draws on retrospectively collected data from 25 individuals with T1D using AID systems. Participants were enrolled during routine endocrinology visits at the Mayo Clinic in Scottsdale, Arizona, between December 2023 and April 2024. Informed consent was obtained in accordance with the approved IRB protocol (IRB \#23-003065).

Each participant contributed real-world data over an average duration of 26 days, resulting in a combined total of 26,707 hours of continuous monitoring. The dataset includes 320,488 entries from CGM devices (Dexcom G6 Pro), along with records of insulin delivery, carbohydrate intake, and device mode classifications (regular, sleep, and exercise), all extracted from the Tandem t:slim X2 insulin pump.

All data were obtained through standard clinical procedures, reflecting typical outpatient care practices. During clinic visits, CGM and pump data were downloaded and archived for clinical use, with no interventions beyond routine monitoring. Eligibility was limited to patients with a confirmed diagnosis of T1D who were actively using both a CGM and a Tandem insulin pump in Control IQ, ensuring the dataset accurately reflects natural treatment behaviors and day-to-day management patterns in real-world settings. Detailed demographic and clinical information is provided in Table~\ref{tab:selected_patients}.

\begin{table}
\centering
\caption{Demographic and Clinical Data of Selected Patients}
\label{tab:selected_patients}
\setlength{\tabcolsep}{2em}
\begin{tabular}{r|clc}
\toprule
No. &  A1c (\%) & Sex &  Age \\
\midrule
         1 &  7.2 &   Male & 65 \\
         2 &  6.6 &   Female & 67 \\
         3 &  5.1 &   Male & 65 \\
         4 &  7.2 &   Female & 69 \\
         5 &  6.5 &   Male & 80 \\
         6 &  6.6 &   Female & 77 \\
         7 &  6.8 &   Male & 36 \\
         8 &  7.2 &   Female & 66 \\
         9 &  8.2 &   Female & 54 \\
        10 &  7.7 &   Female & 71 \\
        11 &  7.3 &   Male & 59 \\
        12 &  6.6 &   Male & 43 \\
        13 &  6.7 &   Male & 80 \\
        14 &  5.0 &   Female & 32 \\
        15 &  5.9 &   Female & 52 \\
        16 &  6.3 &   Male & 40 \\
        17 &  7.1 &   Female & 66 \\
        18 &  6.9 &   Male & 65 \\
        19 &  6.2 &   Male & 27 \\
        20 &  6.7 &   Female & 61 \\
        21 &  6.4 &   Female & 46 \\
        22 &  6.5 &   Female & 46 \\
        23 &  5.7 &   Female & 67 \\
        24 &  6.7 &   Male & 74 \\
        25 &  6.8 &   Male & 72 \\
\bottomrule
\end{tabular}
\end{table}

\begin{figure*}[ht]
    \centering
    \includegraphics[width=0.875\textwidth]{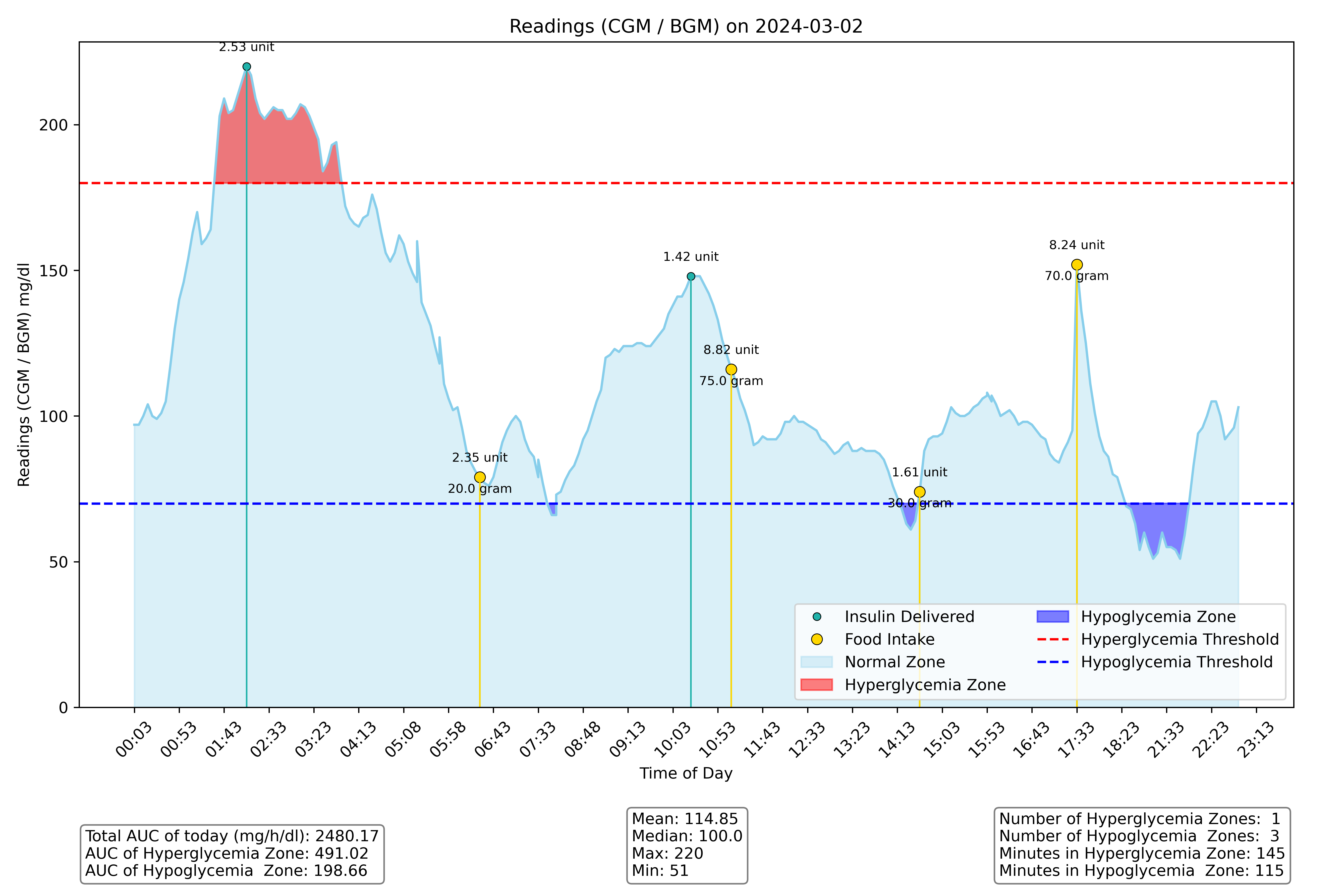}
    \caption{Daily CGM readings with annotated insulin and food events, and providing a summary of glucose trends and variability.}
    \label{fig:statistical_summary}
\end{figure*}

\section{Overview of AZT1D Dataset}
\subsection{Data Pre-processing}
The Tandem insulin pump presents inconsistencies in data availability as it provides data in two formats: i. a CSV file containing blood glucose readings at five-minute interval, carb sizes, target blood glucose levels, bolus logs (including sizes and types) etc., and ii. a PDF with hourly basal rates and device modes (regular/sleep/exercise). Directly using data from these sources is challenging. To bring all data into a common platform, first, timestamps of carb sizes and bolus logs are aligned with blood glucose levels to form a unified timeseries. Second, the basal rates and device modes are extracted from the PDFs by cropping the informative areas and then using an \underline{\textbf{O}}ptical \underline{\textbf{C}}haracter \underline{\textbf{R}}ecognition (OCR) technique. The extraction methods of basal rates and device modes are shown in Fig.~\ref{basal} and \ref{mode}, respectively. 

Since basal insulin is recorded every hour while CGM records are available every 5 minutes, each basal insulin value is repeated across its corresponding 5-minute intervals until the next hourly value. Additionally, a value of 0 is used for carbohydrates and bolus insulin during periods where no records are present.

\subsection{Dataset Description}
The AZT1D dataset consists of multiple records per participant, capturing real-world diabetes management data, including CGM readings, insulin administration events, and carbohydrate intake. Each record contains the following fields:

\begin{itemize}
    \item \textbf{EventDateTime}: The timestamp indicating when the event was recorded.
    \item \textbf{DeviceMode}: The operating mode of the insulin pump at the time of the event, such as regular, sleep, or exercise.
    \item \textbf{BolusType}: The type of bolus insulin delivered, including standard, correction, or automatic boluses.
    \item \textbf{Basal}: The amount of basal insulin (in units) delivered continuously over time.
    \item \textbf{CorrectionDelivered}: The portion of the bolus insulin administered to correct high blood glucose levels.
    \item \textbf{TotalBolusInsulinDelivered}: The total insulin dose (in units) administered during a bolus event.
    \item \textbf{FoodDelivered}: The portion of the bolus insulin dedicated to covering meal carbohydrate intake.
    \item \textbf{CarbSize}: The amount of carbohydrates consumed, measured in grams.
    \item \textbf{CGM}: The recorded glucose value from the continuous glucose monitoring device, measured in mg/dL.
\end{itemize}

This structured dataset enables comprehensive analysis of glycemic trends, insulin dosing strategies, and the impact of lifestyle factors on glucose regulation.

\section{Statistical Analysis}
This section presents foundational analyses conducted at both the individual and population levels, with results shared alongside the dataset to ensure reproducibility. For each day, we generated visualizations showing CGM values, insulin and food intake events, and glycemic excursions. Each figure includes basic statistical metrics such as the mean, max, min, and median. In addition to daily summaries, a monthly report aggregates average values per day to reveal temporal patterns and inter-day relationships. Fig. \ref{fig:statistical_summary} illustrates an example daily summary for one patient.

\begin{figure}[t]
    \centering
    \includegraphics[width=0.475\textwidth]{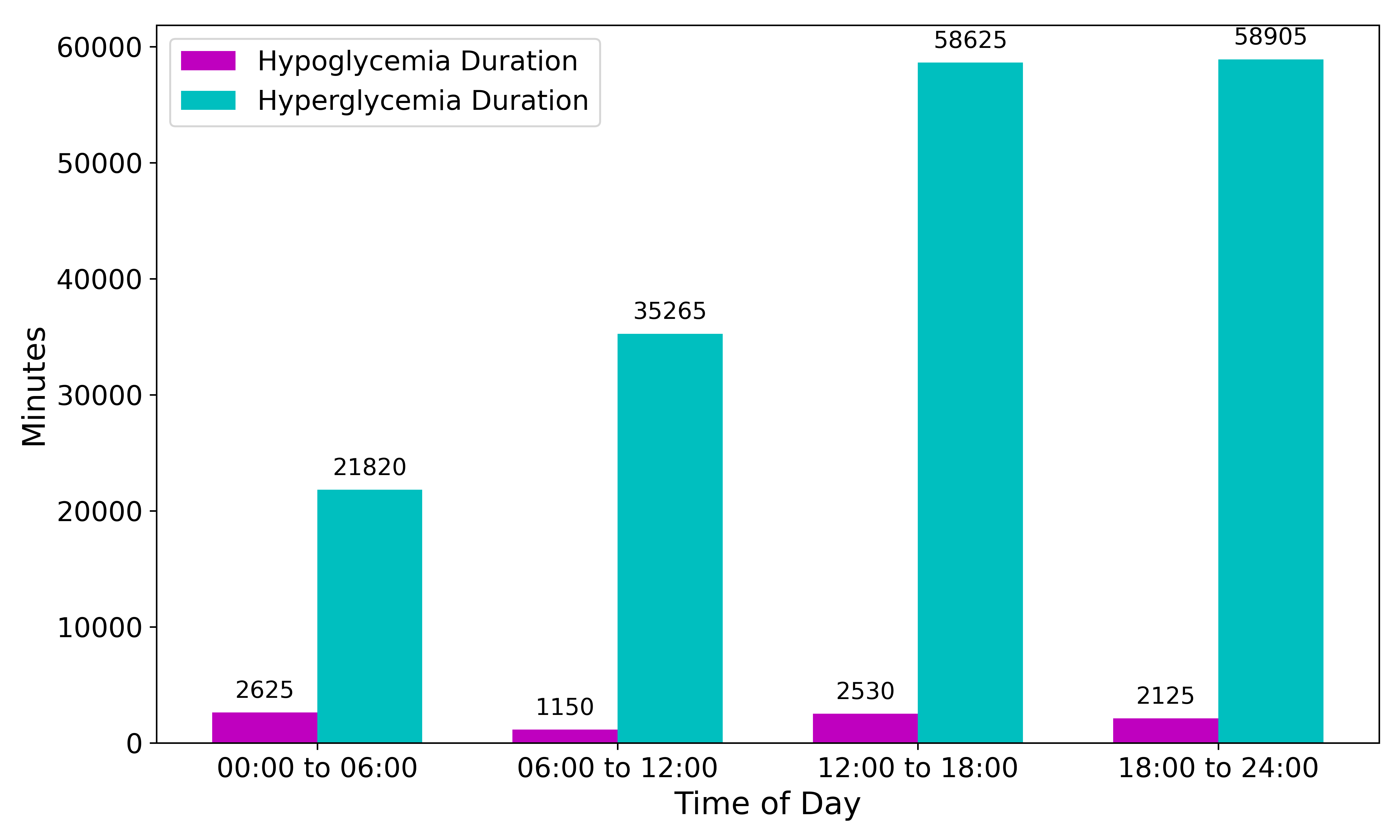}
    \caption{Total duration of hypoglycemia and hyperglycemia events across four time intervals for patients aged 65 and above}
    \label{fig:distribution}
\end{figure}

We analyzed the distribution of glycemic events across four daily intervals in elderly patients ($\ge$ 65 years old), using aggregated data from all patients to capture overall patterns across the population. As shown in Fig.\ref{fig:distribution}, the highest hyperglycemia duration occurred between 18:00 and 24:00 (58,905 minutes), followed by the 12:00 to 18:00 interval. The longest hypoglycemia duration was observed during 00:00 to 06:00 (2,625 minutes), indicating increased vulnerability to low glucose levels during nighttime hours.

In terms of proportional distribution, calculated as the percentage of total recorded time, hyperglycemia accounted for 10.5\% of the time during 00:00 to 06:00, increasing steadily throughout the day to reach 28.6\% in the evening (18:00–24:00). This upward trend likely reflects postprandial glucose rises and possibly fewer corrective actions taken in the evening. In contrast, hypoglycemia remained consistently low, peaking at only 1.3\% in the early morning and remaining below 1\% during the rest of the day. These results, derived from the aggregation of all patient data, emphasize consistent temporal patterns in dysglycemia and suggest the need for time-specific management strategies, particularly enhanced overnight monitoring and afternoon glycemic control for elderly individuals.

\section{Use Cases}
This dataset offers a rich resource for conducting various machine learning and clinical studies in the context of T1D. Its combination of CGM data, insulin delivery logs, carbohydrate intake, and contextual device information provides researchers with the opportunity to explore diverse areas such as blood glucose prediction, counterfactual explanation, temporal data analysis, and the evaluation of AID systems.

For example, one recent study \cite{arefeen2024glyman} introduced a method for generating counterfactual explanations to help patients reduce the risk of hyperglycemia by adjusting their behavior in small but meaningful ways. The proposed framework generated personalized recommendations related to insulin dosing and carbohydrate intake, and incorporated stakeholder preferences to make the interventions more patient-centric. Using this dataset, the method achieved 76.6\% valid explanations and 86\% effectiveness in retrospective evaluations, demonstrating the potential of counterfactual AI for improving glycemic outcomes.

Another study \cite{khamesian2025type} proposed a machine learning model with a custom loss function designed specifically for improving blood glucose predictions in dysglycemic regions—periods when patients are most vulnerable to complications. By leveraging a combination of engineered features and a loss function tuned via a genetic algorithm, the model significantly improved predictive accuracy, reducing RMSE by 23\% and MAE by 31\% on benchmark data. The model’s effectiveness was further validated using this dataset, leveraging its detailed CGM and insulin delivery records. These findings support the utility of the dataset for advancing both predictive modeling and real-world T1D management tools.

Beyond predictive and explanatory models, this dataset also enables the development of offline reinforcement learning frameworks for optimizing insulin dosing. Its temporal structure, detailed bolus information, and CGM data make it well-suited for learning personalized treatment policies from historical records, supporting safer and more adaptive decision-making in T1D management.

\section{Conclusion}
In this work, we introduced AZT1D, a real-world dataset collected from 25 individuals with type 1 diabetes using automated insulin delivery systems. The dataset includes rich contextual information such as CGM readings, insulin delivery events, carbohydrate intake, and device modes over a multi-week period. Through detailed preprocessing and alignment of multiple data sources, AZT1D provides a structured and comprehensive view of diabetes self-management. Initial analyses demonstrated clear temporal patterns in glycemic control among elderly patients and highlighted opportunities for personalized intervention. By making AZT1D publicly available, we aim to support the development of advanced machine learning models, improve glucose forecasting, and foster innovation in personalized diabetes care. To further enhance the dataset’s utility and support broader research applications, we are planning to expand AZT1D to include up to 100 patients as part of our future work.

\section{Acknowledgment}
This work was supported in part by the Mayo Clinic and Arizona State University Seed Grant Program under Award Number ARI-320598. Any opinions, findings, conclusions, or recommendations expressed in this material are those of the authors and do not necessarily reflect the views of the funding organization.

\bibliography{main}

\begin{thebibliography}{1}
\providecommand{\url}[1]{#1}
\csname url@samestyle\endcsname
\providecommand{\newblock}{\relax}
\providecommand{\bibinfo}[2]{#2}
\providecommand{\BIBentrySTDinterwordspacing}{\spaceskip=0pt\relax}
\providecommand{\BIBentryALTinterwordstretchfactor}{4}
\providecommand{\BIBentryALTinterwordspacing}{\spaceskip=\fontdimen2\font plus
\BIBentryALTinterwordstretchfactor\fontdimen3\font minus \fontdimen4\font\relax}
\providecommand{\BIBforeignlanguage}[2]{{%
\expandafter\ifx\csname l@#1\endcsname\relax
\typeout{** WARNING: IEEEtran.bst: No hyphenation pattern has been}%
\typeout{** loaded for the language `#1'. Using the pattern for}%
\typeout{** the default language instead.}%
\else
\language=\csname l@#1\endcsname
\fi
#2}}
\providecommand{\BIBdecl}{\relax}
\BIBdecl

\bibitem{atkinson2014type}
M.~A. Atkinson, G.~S. Eisenbarth, and A.~W. Michels, ``Type 1 diabetes,'' \emph{The lancet}, vol. 383, no. 9911, pp. 69--82, 2014.

\bibitem{diabetes2005intensive}
D.~Control, C.~T. of~Diabetes~Interventions, and C.~D. S.~R. Group, ``Intensive diabetes treatment and cardiovascular disease in patients with type 1 diabetes,'' \emph{New England Journal of Medicine}, vol. 353, no.~25, pp. 2643--2653, 2005.

\bibitem{desalvo2013continuous}
D.~DeSalvo and B.~Buckingham, ``Continuous glucose monitoring: current use and future directions,'' \emph{Current diabetes reports}, vol.~13, pp. 657--662, 2013.

\bibitem{limbert2024automated}
C.~Limbert, A.~J. Kowalski, and T.~P. Danne, ``Automated insulin delivery: A milestone on the road to insulin independence in type 1 diabetes,'' \emph{Diabetes Care}, vol.~47, no.~6, pp. 918--920, 2024.

\bibitem{marling2020ohiot1dm}
C.~Marling and R.~Bunescu, ``The ohiot1dm dataset for blood glucose level prediction: Update 2020,'' in \emph{CEUR workshop proceedings}, vol. 2675, 2020, p.~71.

\bibitem{hidalgo2024hupa}
J.~I. Hidalgo, J.~Alvarado, M.~Botella, A.~Aramendi, J.~M. Velasco, and O.~Garnica, ``Hupa-ucm diabetes dataset,'' \emph{Data in Brief}, vol.~55, p. 110559, 2024.

\bibitem{arefeen2024glyman}
A.~Arefeen, S.~Khamesian, M.~A. Grando, B.~Thompson, and H.~Ghasemzadeh, ``Glyman: Glycemic management using patient-centric counterfactuals,'' in \emph{2024 IEEE EMBS International Conference on Biomedical and Health Informatics (BHI)}.\hskip 1em plus 0.5em minus 0.4em\relax IEEE, 2024, pp. 1--5.

\bibitem{khamesian2025type}
S.~Khamesian, A.~Arefeen, A.~Grando, B.~Thompson, and H.~Ghasemzadeh, ``Type 1 diabetes management using glimmer: Glucose level indicator model with modified error rate,'' \emph{arXiv preprint arXiv:2502.14183}, 2025.

\end{thebibliography}
\bibliographystyle{IEEEtran}

\end{document}